%% file: acl17-hyperpartisan-news-frame.tex
\documentclass[11pt]{article}
\usepackage{acl2015}
\usepackage[american]{babel}
\usepackage[T1]{fontenc}
\usepackage{times}
\usepackage{url}
\usepackage{calc}
\usepackage{microtype}

\newcommand{\bslabel}[1]{\noindent\textsl{#1.}}

\makeatletter
\renewcommand\paragraph{\@startsection{paragraph}{4}{\z@}
  {1.1ex \@plus1ex \@minus.2ex}
  {-1em}%
  {\normalfont\normalsize\bfseries}}
\makeatother

\newif\ifbscomment
\bscommentfalse
\bscommenttrue

\RequirePackage{type1cm}
\RequirePackage{color}
\RequirePackage{soul}
\setstcolor{blue}

\definecolor{upbblue}{rgb}{0.00,0.23,0.47}
\definecolor{darkgray}{gray}{0.40}
\definecolor{mediumgray}{gray}{0.60}
\definecolor{lightgray}{gray}{0.95}
\definecolor{ultralightgray}{gray}{0.98}

\definecolor{upbtextgreen}{rgb}{0.36,0.67,0.15}
\definecolor{upbtextorange}{rgb}{0.88,0.42,0.03}

\usepackage{graphicx}
\usepackage{ifpdf}
\ifpdf
  \DeclareGraphicsExtensions{.pdf,.ai,.jpg,.png}
\else
  \DeclareGraphicsExtensions{.ai,.ps,.eps}
\fi
\graphicspath{{./}{../acl17-hyperpartisan-news-figures/}}

\newcommand{\bsfigure}[3][scale=1.0]{%
  \begin{figure}[tb]
    \centering
    \includegraphics[#1]{#2}
    \vspace{-4ex}
    \caption{#3}\label{#2}
  \end{figure}}

\usepackage{booktabs}
\usepackage{multirow}

\usepackage[flushmargin,hang]{footmisc}

\usepackage{algorithm}
\usepackage[noend]{algpseudocode}
\renewcommand{\textfloatsep}{2ex}

\newcommand{\Ni}{(1)~}
\newcommand{\Nii}{(2)~}
\newcommand{\Niii}{(3)~}

\usepackage{xspace}

\newcommand{\fmeasure}{\ensuremath{F}-Measure\xspace}

\newcommand{\tfidf}{\ensuremath{\mathit{tf}\kern-0.15em\cdot\kern-0.15em\mathit{idf}}}

\begin{document}

\input{acl17-hyperpartisan-news-pre}
\input{acl17-hyperpartisan-news-part1}
\input{acl17-hyperpartisan-news-part2}
\input{acl17-hyperpartisan-news-part3}
\input{acl17-hyperpartisan-news-part4}
\input{acl17-hyperpartisan-news-part5}

\input{acl17-hyperpartisan-news-sum}

\begin{raggedright}
\bibliographystyle{acl}

\input{acl17-hyperpartisan-news-lit}
\end{raggedright}

\end{document}

%% file: acl17-hyperpartisan-news-pre.tex
\title{A Stylometric Inquiry into Hyperpartisan and Fake News}

\author{
Martin Potthast \quad Johannes Kiesel \quad Kevin Reinartz \quad Janek Bevendorff \quad Benno Stein \\[1ex]
Bauhaus-Universität Weimar \\
{\tt $<$first name$>$.$<$last name$>$@uni-weimar.de}
}

\date{}

\maketitle

\begin{abstract}
This paper reports on a writing style analysis of hyperpartisan (i.e., extremely one-sided) news in connection to fake news. It presents a large corpus of 1,627~articles that were manually fact-checked by professional journalists from BuzzFeed. The articles originated from 9~well-known political publishers, 3~each from the mainstream, the hyperpartisan left-wing, and the hyperpartisan right-wing. In sum, the corpus contains 299~fake news, 97\%~of which originated from hyperpartisan publishers.

We propose and demonstrate a new way of assessing style similarity between text categories via Unmasking---a meta-learning approach originally devised for authorship verification---, revealing that the style of left-wing and right-wing news have a lot more in common than any of the two have with the mainstream. Furthermore, we show that hyperpartisan news can be discriminated well by its style from the mainstream ($F_1\!=\!0.78$), as can be satire from both ($F_1\!=\!0.81$). Unsurprisingly, style-based fake news detection does not live up to scratch ($F_1\!=\!0.46$). Nevertheless, the former results are important to implement pre-screening for fake news detectors.
\end{abstract}

%% file: acl17-hyperpartisan-news-part1.tex
\section{Introduction}

The media and the public are currently discussing a new phenomenon called ``fake news'' and its potential role in swaying recent elections, how it may affect democratic societies, and what can and should be done about it. In a nutshell, ``fake news'' encompasses the observation that, in social media, a certain kind of `news' spread much more successfully than others, and that these `news' are typically extremely one-sided (hyperpartisan), inflammatory, emotional, and often riddled with untruths. Although traditional yellow press has been spreading `news' of varying degrees of truthfulness long before the digital revolution, the fact that modern social media amplify fake news to outperform {\em real} news gives many people pause.
The fake news hype caused a widespread disillusionment about social media, and many politicians, news publishers, IT~companies, activists, and scientists concur that this is where to draw the line. For all their good intentions, however, it is already obvious that it must be drawn very carefully (if at all), since nothing less than free speech is at stake---a fundamental right of every free society.

Many favor a two-step approach where first fake news items are detected and then countermeasures are implemented to foreclose false rumors and to discourage repeated offenses. The countermeasures aiming at foreclosing range from displaying warnings, the withholding of news items until stakeholders can react, up to their complete removal. Countermeasures aiming at discouraging range from withholding display advertising revenue, the flagging or downranking of a sender's account, up to banning a sender altogether. By comparison, traditional countermeasures that used to work fairly well in offline media are the publication of refutations, either in a different venue or in the same venue as a fake news item by invoking the so-called {\em right to reply}. In social media, however, the traditional countermeasures are rather ineffective since refutations are typically shared much less than the refuted fake news beforehand, and few jurisdictions have an enforceable right to reply. Once a fake news item spreads virally, the damage is done, and containing it becomes almost impossible: an immediate reaction is crucial.

This paper focuses on fake news detection, but with a twist. While this task is most commonly tackled by automatic fact-checking, we approach it from a different angle by investigating the writing style of fake news in relation to hyperpartisan news. In this regard, we analyze for the first time whether hyperpartisan news can be distinguished by its style from mainstream news (it can), whether satire can be distinguished from both (it can, too), and whether fake news can be detected via style alone (it can't). Furthermore, we introduce a new approach to assess and visualize writing style similarities of text categories based on Unmasking. This way, we assess the style differences between hyperpartisan left-wing news and hyperpartisan right-wing news, showing that the two have significant stylistic similarities. All of these experiments are based on a new, publicly shared dataset, comprising annotations whether news are fake or real, and whether they are hyperpartisan, sampled from 9~well-known publishers and annotated by journalists from BuzzFeed.

After a brief review of related work, Section~\ref{corpus} details the dataset and how it was constructed, Section~\ref{methodology} introduces our methodology including our variant of Unmasking, and Section~\ref{experiments} 
reports the results of the aforementioned experiments.

%% file: acl17-hyperpartisan-news-part2.tex
\section{Related Work}
\label{related-work}

Figure~\ref{taxonomy-fake-news-detection} organizes the literature on fake news detection in terms of three paradigms: fake news detection based on knowledge, on context, and on style. For each of the paradigms we list specific research areas which supply different methods for solving the task. Knowledge-based fake news detection (also called ``fact checking''), is tackled with methods borrowed from information retrieval, semantic web, and linked open data~(LOD) research. Context-based fake news detection employs methods from social network analysis where the spread of false information and rumors as well as their containment is studied. Style-based fake news detection relies on computational linguistics and natural language processing, and, more specifically, on methods from deception detection to identify statements at the sentence-level that constitute falsehoods and lies. One field of research that has hardly been considered in the context of fake news detection, yet, is style-based text categorization. In this paper, we will close this gap and analyze the potential of classifying news items by the style of their text body into classes corresponding to fake, real, and satire, as well as to hyperpartisan and mainstream news.

\bsfigure[width=\columnwidth]{taxonomy-fake-news-detection}{Taxonomy of paradigms for fake news detection alongside a selection of relevant work.}

\medskip
\bslabel{Knowledge-based Fake News Detection}
Methods from information retrieval have been proposed early on to determine the truthfulness of a given web resource. For example, \newcite{etzioni:2008} propose to use their well-known tool Text Runner \cite{yates:2007} to extract and index factual knowledge from the web, and to use the same technology to extract factual statements from a given text in question, matching them against the indexed facts to identify inconsistencies. \newcite{magdy:2010} develop a statistical model to check factual statements extracted from a given document in question, analyzing how frequently they are supported by documents retrieved from the web. Both approaches presume that web resources (or the frequency by which a fact is mentioned) can be used as an indication of its truth. However, the problem of this argument is that the reputation and reliability of almost any website can be put into question. In this regard, \newcite{ginsca:2015} give a comprehensive overview of challenges and approaches from the literature to assess the various aspects of credibility in information retrieval, namely: expertise, trustworthiness, quality, and reliability.

While the extraction and retrieval of factual information plays an important role when using the web as a knowledge base, knowledge becomes increasingly shared as structured knowledge bases, integrated in the linked open data cloud, inducing the semantic web. In fact, information extracted from the web, too, is typically stored in such knowledge bases. When presuming that factual knowledge is available in sufficient detail for a domain of interest, the task of fake news detection boils down to checking whether a given fact is already known, or whether it can be inferred from other facts. \newcite{wu:2014} try to assess the truthfulness of a given fact by ``perturbing'' it, formulating queries to knowledge bases and interpreting the variation in the results as a sign of whether the fact is either strongly or weakly supported. \newcite{ciampaglia:2015} cast fact-checking as a problem of finding shortest paths between concepts in a knowledge graph; they propose a metric to assess the truth of a statement by analyzing path lengths between the concepts in question. Conversely, \newcite{shi:2016} cast fake news detection as a link prediction task, where a probability is estimated in order to decide whether concepts covered by a to-be-checked statement should be linked.

\medskip
\bslabel{Context-based Fake News Detection}
By considering the mechanisms of social networks, new angles on the problem of fake news dissemination come into reach. \newcite{acemoglu:2010} model how (mis-)information is spread in social networks, and \newcite{budak:2011} and \newcite{nguyen:2012} propose algorithms to limit their spread. \newcite{kwon:2013} combine social network analysis and linguistic feature obtained from applying LIWC \cite{pennebaker:2003} to identify rumors as they spread. Studying the spread of misinformation on Facebook during an election, \newcite{mocanu:2015} provide evidence that unsubstantiated claims spread as widely as well-established ones, and that user groups with a predisposition to conspiracy theories are more open to sharing misinformation. \newcite{tambuscio:2015} also study the spread of misinformation in social media; however, they also study the efficacy of countermeasures such as debunking sites. In particular, they find that by exceeding a certain threshold in spreading the refutation is sufficient to remove the misinformation from the network, and that this threshold does not depend on the spreading rate but on credulity and forgetfulness.

\medskip
\bslabel{Style-based Fake News Detection}
Another approach to fake news detection is to sidestep fact checking and social network analysis altogether by modeling the {\em nature of faking} and its manifestation in text. Two branches of research provide the rationale and methodology: deception detection in text and style-based text categorization. Deception detection originates from forensic linguistics and builds on the Undeutsch hypothesis---a result from forensic psychology asserting that memories of real-life, self-experienced events differ in content and quality from imagined events \cite{undeutsch:1967}. The hypothesis leads to the development of (the now commonly applied) forensic tools to assess testimony at the statement level, such as Criteria-based Content Analysis~(CBCA) and Scientific Content Analysis~(SCAN). Nowadays, technology is being developed to operationalize deception detection at scale, e.g. in order to identify or to detect uncertainty in social media posts: \newcite{wei:2013} propose a model to detect tweets that convey uncertain information. Regarding fake news detection, \newcite{chen:2015b} point out the need for an ``automatic crap detector'' for news, but do not report on actual experiments, whereas \newcite{rubin:2015a} apply, for the first time, deception detection approaches to fake news detection using rhetorical structure theory as a measure of story coherence.

Style-based text categorization was proposed by \newcite{argamon:1998} as an alternative to topic-based text categorization in order to tackle tasks ranging from author profiling (by age, gender, native language, etc.) to broader style categories such as text genre. Since it is hypothesized that deception has its own style as well, there is an overlap in methodology, using full text classification as another means to assess the truthfulness of a given text as opposed to analyzing individual statements. For example, \newcite{afroz:2012} attempt to detect texts whose authors tried to obfuscate their writing style to deflect author identification. As an early precursor to fake news detection, \newcite{badaskar:2008} train models to tell real news apart from news that have been automatically generated using a language model, However, \newcite{rubin:2016} contributed the first actual attempt at fake news detection by separating satire news as a representative of humorous fakes from real news in a dataset of 180~news articles each, achieving \fmeasure values between~0.82 and~0.87 for various variants of a \tfidf-weighted lexical vector space model. We employ this dataset in conjunction with our own in our experiments to study the connection of fake news, real news, and satire for the first time.

%% file: acl17-hyperpartisan-news-part3.tex
\section{The BuzzFeed-Webis Fake News Corpus 2016}
\label{corpus}

This section introduces the BuzzFeed-Webis Fake News Corpus~2016, detailing its construction and annotation by professional journalists employed at BuzzFeed, as well as key figures and statistics.

\subsection{Corpus Construction}

The corpus comprises a complete sample of the output of 9~publishers in a week close to the US~elections. Among the selected publishers are 6~prolific hyperpartisan publishers (three left-wing and three right-wing) and three mainstream publishers (see Table~\ref{table-corpora}). All publishers earned Facebook's blue checkmark~\includegraphics[height=9pt]{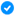}, indicating authenticity and an elevated status within the network. For seven weekdays (September~19 to~23 and September~26 and~27), every post and linked news article of the 9~publishers was fact-checked claim-by-claim by 5~BuzzFeed journalists, including about~10\% of posts forwarded from third parties. \newcite{silverman:2016} reported key insights as a data journalism article, having checked a total of 2,282~posts, 1,145~of which from mainstream publishers, 471~from hyperpartisan left-wing publishers, and 666~from hyperpartisan right-wing publishers. Alongside the article, the annotations were published as well.%
\footnote{\fontsize{8.4pt}{9.4pt}\selectfont http://github.com/BuzzFeedNews/2016-10-facebook-fact-check}
However, this data only comprises URLs to the original Facebook posts. To create the corpus, we hence archived the posts, the linked articles, and attached media as well as relevant meta data to ensure long-term availability. Due to the rapid pace at which the publishers change their websites, we were able to recover only 1,627~articles, 826~mainstream, 256~left-wing, and 545~right-wing. Table~\ref{table-corpora} gives an overview.

\medskip
\bslabel{Manual Fact-checking}
Five BuzzFeed journalists conducted the manual fact-checks of the news articles. To avoid bias with regard to publishers, news from all publishers and all days were assigned round robin. It became clear very quickly that a binary distinction between fake and real news was infeasible, since hardly any piece of fake news is entirely false, and hardly any piece of real news is flawless. Therefore, posts were rated ``mostly true,'' ``mixture of true and false,'' ``mostly false,'' or, if the post was opinion-driven or otherwise lacked a factual claim, ``no factual content.'' The ratings ``mixture of true and false'' and ``mostly false'' had to be justified, and, when in doubt about a rating, a second opinion was collected, whereas disagreements were resolved by a third one. Finally, all news rated ``mostly false'' underwent a final check by a different rater, to ensure the rating was justified. Raters were given the following guidance:

\input{table-corpora}

Mostly true: the post and any related link or image are based on factual information and portray it accurately. This lets the authors interpret the event/info in their own way, so long as they do not misrepresent events, numbers, quotes, reactions, etc., or make information up. This rating does not allow for unsupported speculation or claims.

Mixture of True and False (mix, for short): Some elements of the information are factually accurate, but some elements or claims are not. This rating should be used when speculation or unfounded claims are mixed with real events, numbers, quotes, etc., or when the headline of the link being shared makes a false claim but the text of the story is largely accurate. It should also only be used when the unsupported or false information is roughly equal to the accurate information in the post or link. Finally, use this rating for news articles that are based on unconfirmed information.

Mostly False: Most or all of the information in the post or in the link being shared is inaccurate. This should also be used when the central claim being made is false.

No Factual Content (n/a, for short): This rating is used for posts that are pure opinion, comics, satire, or any other posts that do not make a factual claim. This is also the category to use for posts that are of the ``Like this if you think...'' variety.

\subsection{Corpus Statistics}

Table~\ref{table-corpora} shows the fact-checking results and some key statistics per article. Unsurprisingly, none of the mainstream articles are mostly false, whereas 8~across all three publishers are a mixture of true and false. Disregarding non-factual articles, a little more than a quarter of all hyperpartisan left-wing articles were found faulty: 15~articles mostly false, and~51 a mixture of true and false. Publisher ``The Other 98\%'' sticks out by achieving an almost perfect score. By contrast, almost 45\% of the right-wing articles are a mixture of true and false~(153) or mostly false~(72). Here, publisher ``Right Wing News'' sticks out by supplying more than half of mixtures of true and false alone, whereas mostly false articles are equally distributed.

Regarding key statistics per article, it is interesting to note that the articles from all mainstream publishers are on average about 20~paragraphs long with word counts ranging from 550~words on average at ABC News to~800 at Politico. Except for one publisher, left-wing articles and right-wing articles are shorter on average in terms of paragraphs as well as word count, averaging at about 420 words and 400 words, respectively. Left-wing articles quote on average about 10~words more than the mainstream, and right-wing articles 6~words. When articles comprise links, they are usually external ones, whereas ABC News rather uses internal links, and only half of the links found at Politico articles are external. Left-wing news articles stick out by containing almost double the amount of links across publishers than mainstream and right-wing news.

\subsection{Operationalizing Fake News}

In our experiments, we operationalize the category of fake news by joining the articles that were rated mostly false with those rated a mixture of false and true. Arguably, the latter may not be exactly what is colloquially understood under the term ``fake news'' (as in: a complete fabrication), however, practice shows fake news are hardly ever devoid of truth. More often, true facts are misconstrued using argumentative fallacies to influence a person's opinion. In our experiments, we hence call mostly true articles real news, mostly false plus mixtures of true and false, except for satire, fake news, and disregard all articles rated non-factual.

%% file: table-corpora.tex
\begin{table}[t!]%
\centering%
\scriptsize%
\setlength{\tabcolsep}{2pt}%
\vspace{-2ex}%
\caption{The BuzzFeed-Webis Fake News Corpus 2016 at a glance. (``Paras.'' short for ``paragraphs'')}%
\label{table-corpora}%
\vspace{1ex}%
\begin{tabular}{@{}l@{\kern-0.5em}rrcrr@{\hspace{12pt}}r@{\hspace{5pt}}rr@{\hspace{5pt}}rr@{}}
\toprule
\bfseries\em Orientation & \multicolumn{5}{@{}c@{\hspace{15pt}}}{\bfseries Fact-checking results} & \multicolumn{5}{@{}c@{}}{\bfseries Key statistics per article} \\
\cmidrule(l@{\tabcolsep}r@{12pt}){2-6}
\cmidrule(l@{-0.25em}r){7-11}
\smash{\raisebox{5pt}{Publisher}} & true\, & \multicolumn{1}{@{}c@{\kern-0.15em}}{mix} & false & \multicolumn{1}{@{}c@{\kern-0.15em}}{n/a} & $\Sigma$ \ \ & \multicolumn{1}{@{\kern-0.25em}l@{}}{Paras.} & \multicolumn{2}{@{\kern-0.3em}c@{\hspace{6pt}}}{Links} & \multicolumn{2}{@{}c@{}}{Words} \\
\cmidrule(l@{-0.4em}r@{6pt}){8-9}
\cmidrule(l@{1pt}r){10-11}
&&&&&&& \multicolumn{1}{@{\kern-0.3em}c@{\kern-0.1em}}{extern} & all\, & \multicolumn{1}{@{}c@{\kern-0.25em}}{quoted} & all \ \ \ \\
\midrule
\em Mainstream   &  806 &   8 & \phantom{0}0 & 12 & 826 & 20.1 & 2.2 & 3.7 & 18.1 & 692.0 \\
ABC News         &   90 &   2 & \phantom{0}0 & 3 & 95 & 21.1 & 1.0 & 4.8 & 21.0 & 551.9 \\
CNN              &  295 &   4 & \phantom{0}0 & 8 & 307 & 19.3 & 2.4 & 2.5 & 15.3 & 588.3 \\
Politico         &  421 &   2 & \phantom{0}0 & 1 & 424 & 20.5 & 2.3 & 4.3 & 19.9 & 798.5 \\
\midrule
\em Left-wing    &  182 &  51 &           15 & 8 & 256 & 14.6 & 4.5 & 4.9 & 28.6 & 423.2 \\
Addicting Info   &   95 &  25 & \phantom{0}8 & 7 & 135 & 15.9 & 4.4 & 4.5 & 30.5 & 430.5 \\
Occupy Democrats &   59 &  25 & \phantom{0}7 & 0 & 91 & 10.9 & 4.1 & 4.7 & 29.0 & 421.7 \\
The Other 98\%   &   28 &   1 & \phantom{0}0 & 1 & 30 & 20.2 & 6.4 & 7.2 & 21.2 & 394.5 \\
\midrule
\em Right-wing   &  276 & 153 &           72 & 44 & 545 & 14.1 & 2.5 & 3.1 & 24.6 & 397.4 \\
Eagle Rising     &  106 &  47 &           25 & 36 & 214 & 12.9 & 2.6 & 2.8 & 17.3 & 388.3 \\
Freedom Daily    &   49 &  24 &           22 & 4 & 99 & 14.6 & 2.2 & 2.3 & 23.5 & 419.3 \\
Right Wing News  &  121 &  82 &           25 & 4 & 232 & 15.0 & 2.5 & 3.6 & 33.6 & 396.6 \\
\midrule
$\Sigma$         & 1264 & 212 &           87 & 64 & 1627 & 17.2 & 2.7 & 3.7 & 20.6 & 551.0 \\
\bottomrule
\end{tabular}%
\end{table}

%% file: acl17-hyperpartisan-news-part4.tex
\section{Methodology}
\label{methodology}

In this section, we briefly review methodology, including a brief recap of Unmasking by \newcite{koppel:2007}, for which we investigate for the first time its use in distinguishing genre styles as opposed to authors, and our set of features used to capture writing style. For sake of reproducibility, all our code will be made publicly available.

\subsection{Unmasking Style Categories}

Unmasking, as proposed by \newcite{koppel:2007}, is a meta learning approach that was originally intended for authorship verification. In this paper, we study for the first time whether it can be used to assess the similarity of more broadly defined style categories compared to authorial style, such as left-wing versus right-wing versus mainstream news. This way, we attempt uncover relations between the writing styles that people may involuntarily adopt as per their political orientation.

Originally, Unmasking takes two documents as input and outputs its confidence whether they have been written by the same author. Three steps are taken to accomplish this: first, each document is chunked into a set of at least 500-word long chunks; second, reconstruction errors are measured while iteratively removing the most discriminative features of a style model comprising the 250 most frequent words used to separate the two chunk sets with a linear classifier; and third, the resulting reconstruction error curves are analyzed with regard to their slope. A steep decrease is more likely than a shallow decrease if the two documents have been written by the same author, since there are presumably less discriminating features between documents written by the same author than between documents written by different authors. Training a classifier on many examples of error curves obtained from same-author document pairs and different-author document pairs yields an effective authorship verifier---at least for long documents that can be split up into a sufficient number of chunks.

We believe that what applies to the style of authors also applies to more broadly defined styles; in our case hyperpartisanship. We adapt Unmasking by skipping its first step and using two sets of documents (e.g., left-wing articles and right-wing articles) as input. Further, we plot the reconstruction error curves for visual inspection: steeper decreases in these plots indicate style similarity of the two input document sets, just as they do when using chunks obtained from documents written by the same author. This way, we demonstrate that Unmasking can be applied in situations where the question arises whether man-made categories of texts are stylistically discriminative, or not. When applied to document sets sampled from hyperpartisan left, hyperpartisan right, and mainstream publishers, we expect insights into the nature of the writing style unconsciously adopted by people of these orientations.

\subsection{Style Features and Feature Selection}

Our writing style model incorporates commonly used style features as well as some specific to the news domain. The former are n-grams of characters, stop words (in order of appearance in the text), and parts-of-speech with~n in~$[1,3]$. Furthermore, we employ 10~readability scores%
\footnote{\fontsize{8.6pt}{9.6pt}\selectfont%
Automated Readability Index,
Coleman Liau Index,
Flesh Kincaid Grade Level and Reading Ease,
Gunning Fog Index,
LIX,
McAlpine EFLAW Score,
RIX,
SMOG Grade,
Strain Index}
as well as dictionary features, where each one indicates the frequency of words from a tailor-made dictionary in a given document, using the General Inquirer Dictionaries as a basis \cite{stone:1966}. The domain-specific features include ratios of quoted words and external links, and the number of paragraphs and their average length in a document.

In each of our experiments, we carefully select from the aforementioned features the ones worthwhile using. To avoid overfitting, all features that are hardly represented in the documents of our corpus (i.e., occur in less than 10\% of documents) are discarded. This pertains particularly to many individual n-grams as well as some dictionary features, since most low-frequency n-grams found are unique to a certain publisher. We further disregard features that are not represented in at least two of the categories to be distinguished in a given experiment. This way, the performance of our classifiers may not be as high as it could be when using all features, but our results can be used to draw meaningful conclusions.

\subsection{Baselines}

\enlargethispage{\baselineskip}
We employ the standard bag of word model for a topic-based classification baseline. This approach is less practical, however, since topics change frequently and drastically in the news domain. Moreover, we supply naive baselines that classify all items into one of the classes in question, thus relating results to the class distributions.

%% file: acl17-hyperpartisan-news-part5.tex
\section{Experiments}
\label{experiments}

We report on the results of two series of experiments aimed at investigating the style differences and similarities between hyperpartisanship and the mainstream, as well as between fake, real, and satire news. In particular, we shed light on the following three questions:
\Ni
\emph{Is there a common style of hyperpartisanship?}
Our working hypothesis is that left-wing and right-wing hyperpartisans have more in common than they themselves would admit, and our results indeed provide first evidence in the affirmative.
\Nii
\emph{Is style-based fake news detection feasible?}
Of course, we do not expect to solve fake news detection using style alone, but exploiting style has the advantage that it can be applied for pre-screening in real time, and that authors presumably have little control over their own writing style at large.
\Niii
\emph{Can hyperpartisan fake news be distinguished from satire?}
Investigating the special case of satire news is important, as we cannot allow for humor to be sacrificed on the altar of truth. Therefore, any fake news detection approach must be able to deal with satire.

\subsection{Hyperpartisanship vs.\ Mainstream}
\label{hyperpartisanship-vs-mainstream}

This series of experiments targets the political orientation, and thus research question~(1). We conduct three experiments, where the first two distinguish left-wing, right-wing, and mainstream, and the third one hyperpartisan and mainstream.

\medskip
\bslabel{A.~Unmasking hyperpartisanship}
We apply Unmasking as described above onto pairs of the three orientations in question. Figure~\ref{plot-unmasking-orientation} shows the resulting Unmasking curves (Unmasking is symmetrical, hence three curves). The curves are averaged over 5~runs, where each run comprised sets of 100~documents from each orientation in question. In case of the left-wing orientation, where less than 500~documents are available in our corpus, once all of them had been used, they were shuffled again to select documents for the remainder of the runs.

\bsfigure[width=\columnwidth]{plot-unmasking-orientation}{Unmasking applied to pairs of political orientations. The quicker a curve decreases, the more similar the respective styles are.}

As all three curves decrease quickly, Koppel et~al.'s original hypothesis in the context of authorship verification would be that all articles have been written by the same author. When transferred to our application of Unmasking, one would generally conclude that all pairs of orientations have a common style, which is unsurprising given that the documents have all been sampled from news publishers. They all possess the style of the genre of news articles. However, Unmasking allows for a more fine-grained assessment of style similarity. In this regard, we formulate the Unmasking hypothesis as follows: style similarity is characterized by the slope of a given Unmasking curve, where a steeper decrease indicates higher similarity. Marked differences between the curves originating from different pairs of orientations can be observed. If the Unmasking hypothesis is correct, this experiment shows that the left-wing and right-wing documents found in our corpus have a lot more in common, stylistically, than documents from either orientation with mainstream documents.

\medskip
\bslabel{B.~Predicting Hyperpartisanship}
Given this initial result, we follow that it should be possible to discriminate hyperpartisan news from the mainstream, even if one has only examples from the respective other hyperpartisan side plus mainstream examples. To test this hypothesis, we train three binary classifiers (random forests) to distinguish hyperpartisan news from mainstream news, while omitting left-wing documents from the training set, omitting right-wing articles from the training set, and retaining both, respectively. The training sets for all three classifiers were balanced between the classes mainstream and hyperpartisan; where left-wing or right-wing documents were omitted, oversampling was applied on the remaining documents from the respective other side. 

\input{table-hyperpartisanship}

Table~\ref{table-hyperpartisanship} shows performance values as averages of 3-fold cross-validation, where each fold comprises one publisher from each orientation so that the classifier can not learn the publisher style. Besides our style feature model, we also provide performance values of corresponding classifiers trained under a topic feature model (bag of words) as baseline for comparison. When omitting left-wing documents from the training set, the style-based classifier still achieves 0.74~accuracy on left-wing test documents. Likewise, it achieves 0.66~accuracy classifying right-wing documents without having trained on them. In both cases, the style-based classifier outperforms the baseline topic feature model. This behavior supports our observation that the style of left-wing documents is not so different from that of right-wing ones. Unsurprisingly, not omitting any documents from the training data improves accuracy significantly, however, only for the class of hyperpartisan documents. The mainstream class can be best discriminated when omitting right-wing documents, at the expense of accuracy for the class of hyperpartisan documents. 

Table~\ref{table-hyperpartisan-mainstream} shows performance values of the binary classifier, trained with both hyperpartisan sides present in its training set to discriminate hyperpartisan and mainstream news. The best classification accuracy of~0.75 at a remarkable 0.89~recall for the hyperpartisan class is achieved by the style-based classifier, outperforming the topic feature model.

\input{table-hyperpartisan-mainstream}

\medskip
\bslabel{C.~Predicting Orientation}
As a last experiment in this series, we trained a classifier to predict the orientation of an individual news article. Table~\ref{table-orientation} shows the performance values of a random forest in terms of accuracy, precision, and recall, again, as averages of 3-fold cross-validation, compared to the same baselines. Interestingly, in the three-class setting, the topic baseline outperforms the style-based model with regard to accuracy, whereas the results for class-wise precision and recall are a mixed bag. The left-wing documents are apparently significantly more difficult to be identified compared to documents from the other two orientations. When looking at the confusion matrix, it turns out that 66\%~of misclassification of left-wing documents are falsely classified as right-wing documents. The reverse, however, is not true: 60\%~of all misclassified right-wing documents are classified as mainstream documents, whereas misclassified mainstream documents spread almost equally across the other classes.

\input{table-orientation}

\subsection{Fake vs.\ Real (vs.\ Satire)}

This series of experiments targets research questions~(2) and~(3). Again, we conduct three experiments, where the first is about predicting veracity, and the last two about discriminating satire.

\medskip
\bslabel{A.~Predicting veracity}
Given the encouraging results in predicting hyperpartisan news---in particular the  high recall of~0.89 at a reasonable precision of~0.69---we are confident that, with some further effort, a practical classifier can be built that detects hyperpartisan news at scale. When taking into account that the mainstream news publishers in our corpus did not publish any news that are mostly false, and only very few instances that are mixtures of true and false, we can safely disregard them for the task of fake news detection. In this connection, a classifier that reliably distinguishes hyperpartisan news from the mainstream can act as a pre-filter for a subsequent, more in-depth fake news detection approach, which may in turn be tailored to a much more narrowly defined classification task. We hence use only the left-wing documents and the right-wing documents of our corpus as a basis for training our fake news classifier.

\input{table-fake}

Table~\ref{table-fake} shows the performance values for three classifiers, a generic classifier that predicts fake news across sides, and orientation-specific classifiers that have been individually trained on documents from either side, left-wing and right-wing news. In both cases, values are again averages of 3-fold cross-validation. Although all classifiers outperform the naive baselines of classifying everything as one class in terms of precision, the slight increase comes at the cost of a large decrease in recall. While the orientation-specific classifiers are slightly better for most metrics, none of them can outperform the naive baselines with regard to the \fmeasure.

\bsfigure[width=\columnwidth]{plot-unmasking-satire}{Unmasking applied to pairs of sets of news that are fake, real, and satire.}

\medskip
\bslabel{B.~Unmasking satire}
As mentioned above, a special kind of harmless fake news is satire, which takes the form of news but lies more or less obviously to amuse its readers. Regardless the problems that fake news spread may cause, satire and other forms of humor should never be filtered. We assess the style similarity of satire compared to fake news and real news for the firs time, again applying Unmasking to compare pairs of the three categories of news as described above. For satire news, we use the 180~articles from the S-n-L News DB~(\newcite{rubin:2016}, Section~\ref{related-work}). Figure~\ref{plot-unmasking-satire} shows the resulting Unmasking curves. The slope for the pair of fake vs real news drops somewhat slower compared to the other two pairs. Apparently, the style of fake news has much more in common with that of real news than either of the two has with satire (cf.\ Section~\ref{hyperpartisanship-vs-mainstream}~A). These results are encouraging: satire is apparently not very similar to either fake news or real news, so that, just like with hyperpartisan news compared to mainstream news, it should be fairly easy to discriminate between the two. This is indeed the case, as our last experiment confirms.

\medskip
\bslabel{C.~Discriminating satire}
Finally, we evaluate the use of style features for predicting satire. Table~\ref{table-satire} shows the performance values of our random forest classifier in the satire-detection setting used by \newcite{rubin:2016}. This setting uses a balanced 3:1 training-to-test set split over 360~articles (180 per class). As can be seen, our style-based model significantly outperforms all baselines across the board, achieving an accuracy of~0.82, and an $F$~score of~0.81. While it does not outperform the classifier based on topic, absurdity, grammar, and punctuation features by \newcite{rubin:2016}, it clearly improves over a pure topic classification. We argue that incorporating topic into satire detection is not appropriate, since the topics of satire change as the topics of news in general do. A classifier with topic features would therefore not generalize. Apparently, a style-based model is competitive, and we believe that satire can be detected at scale this way, so as to prevent other fake news detection technology from falsely filtering it.

\input{table-satire}

%% file: table-hyperpartisanship.tex
\begin{table}[t!]%
\vspace{-2ex}%
\caption{Accuracies of predicting hyperpartisanship vs mainstream under different training sets.}%
\label{table-hyperpartisanship}%
\centering%
\scriptsize%
\setlength{\tabcolsep}{3.9pt}%
\begin{tabular}{@{}l@{\hspace{10pt}}ccc@{\hspace{10pt}}ccc@{\hspace{10pt}}ccc@{}}
\toprule
\bfseries Features & \multicolumn{9}{c}{\bfseries Training set} \\
\cmidrule{2-10}
& \multicolumn{3}{@{}c@{\hspace{10pt}}}{\bfseries without left} & \multicolumn{3}{@{}c@{\hspace{10pt}}}{\bfseries without right} & \multicolumn{3}{@{}c@{}}{\bfseries with both} \\
\cmidrule(r{10pt}){2-4}
\cmidrule(r{10pt}){5-7}
\cmidrule(){8-10}
\hfill Test set:\kern-1em & left & \kern-0.1em right & \kern-0.25em main. & left & \kern-0.1em right & \kern-0.25em main. & left & \kern-0.1em right & \kern-0.25em main. \\
\midrule
Style & \bfseries 0.74 & 0.84 & 0.66 & 0.68 & \bfseries 0.66 & 0.74 & 0.90 & 0.89 & 0.62 \\
Topic & 0.68 & 0.77 & 0.78 & 0.50 & 0.48 & 0.87 & 0.79 & 0.85 & 0.60 \\
\bottomrule
\end{tabular}%
\end{table}

%% file: table-hyperpartisan-mainstream.tex
\begin{table}[t!]%
\vspace{-2ex}%
\caption{Accuracies, class-wise precision, recall, and \fmeasure of predicting hyperpartisanship. Including baselines predicting all articles as hyperpartisan or mainstream, respectively}%
\label{table-hyperpartisan-mainstream}%
\scriptsize%
\centering%
\setlength{\tabcolsep}{5.25pt}%
\begin{tabular}{@{}l@{\hspace{8pt}}c@{\hspace{15pt}}cc@{\hspace{15pt}}cc@{\hspace{15pt}}cc@{}}
\toprule
\bfseries Features & \bfseries Accuracy & \multicolumn{2}{@{}c@{\hspace{15pt}}}{\bfseries Precision} & \multicolumn{2}{@{}c@{\hspace{15pt}}}{\bfseries Recall} & \multicolumn{2}{@{}c@{}}{\bfseries F$_{\mathbf{1}}$} \\
\cmidrule(r{15pt}){2-2}
\cmidrule(r{15pt}){3-4}
\cmidrule(r{15pt}){5-6}
\cmidrule{7-8}
& all & hyp. & \kern-0.25em main. & hyp. & \kern-0.25em main. & hyp. & \kern-0.25em main. \\
\midrule
Style & 0.75 & 0.69 & 0.86 & 0.89 & 0.62 & 0.78 & 0.72 \\
Topic & 0.71 & 0.66 & 0.79 & 0.83 & 0.60 & 0.74 & 0.68 \\
\midrule
All-hyp. & 0.49 & 0.49 & - & 1.00 & 0.0 & 0.66 & - \\
All-main. & 0.51 & - & 0.51 & 0.0 & 1.00 & - & 0.68 \\
\bottomrule
\end{tabular}%
\end{table}

%% file: table-orientation.tex
\begin{table}[t!]%
\vspace{-2ex}%
\caption{Accuracies and class-wise precision, recall and \fmeasure of predicting orientation.}%
\label{table-orientation}%
\scriptsize%
\centering%
\setlength{\tabcolsep}{2pt}%
\begin{tabular}{@{}l@{\hspace{5pt}}c@{\hspace{6pt}}ccc@{\hspace{6pt}}ccc@{\hspace{6pt}}ccc@{}}
\toprule
\bfseries Features & \bfseries Accuracy & \multicolumn{3}{@{}c@{\hspace{6pt}}}{\bfseries Precision} & \multicolumn{3}{@{}c@{\hspace{6pt}}}{\bfseries Recall} & \multicolumn{3}{@{}c@{}}{\bfseries F$_{\mathbf{1}}$} \\
\cmidrule(r{6pt}){2-2}
\cmidrule(r{6pt}){3-5}
\cmidrule(r{6pt}){6-8}
\cmidrule{9-11}
& all & left & \kern-0.1em right & \kern-0.25em main. & left & \kern-0.1em right & \kern-0.25em main. & left & \kern-0.1em right & \kern-0.25em main. \\
\midrule
Style & 0.60 & 0.21 & 0.56 & 0.75 & 0.20 & 0.59 & 0.74 & 0.20 & 0.57 & 0.75 \\
Topic & 0.64 & 0.24 & 0.62 & 0.72 & 0.15 & 0.54 & 0.86 & 0.19 & 0.58 & 0.79 \\
\midrule
All-left & 0.16 & 0.16 & - & - & 1.00 & 0.0 & 0.0 & 0.27 & - & - \\
All-right & 0.33 & - & 0.33 & - & 0.0 & 1.00 & 0.0 & - & 0.50 & - \\
All-main. & 0.51 & - & - & 0.51 & 0.0 & 0.0 & 1.00 & - & - & 0.68 \\
\bottomrule
\end{tabular}%
\end{table}

%% file: table-fake.tex
\begin{table}[t!]%
\vspace{-2ex}%
\caption{Accuracies, class-wise precision, recall and \fmeasure of predicting fake news by features and classifier.}
\label{table-fake}%
\scriptsize%
\centering%
\setlength{\tabcolsep}{5pt}%
\begin{tabular}{@{}l@{\hspace{15pt}}c@{\hspace{15pt}}cc@{\hspace{15pt}}cc@{\hspace{15pt}}cc@{}}
\toprule
\bfseries Features & \bfseries Accuracy & \multicolumn{2}{@{}c@{\hspace{15pt}}}{\bfseries Precision} & \multicolumn{2}{@{}c@{\hspace{15pt}}}{\bfseries Recall} & \multicolumn{2}{@{}c@{}}{\bfseries F$_{\mathbf{1}}$} \\
\cmidrule(r{15pt}){2-2}
\cmidrule(r{15pt}){3-4}
\cmidrule(r{15pt}){5-6}
\cmidrule{7-8}
& all & fake & real & fake & real & fake & real \\
\midrule
\multicolumn{7}{@{}l@{}}{\it Generic classifier} \\
Style & 0.55 & 0.42 & 0.62 & 0.41 & 0.64 & 0.41 & 0.63 \\
Topic & 0.52 & 0.41 & 0.62 & 0.48 & 0.55 & 0.44 & 0.58 \\
\midrule
\multicolumn{7}{@{}l@{}}{\it Orientation-specific classifier} \\
Style & 0.55 & 0.43 & 0.64 & 0.49 & 0.59 & 0.46 & 0.61 \\
Topic & 0.58 & 0.46 & 0.65 & 0.45 & 0.66 & 0.46 & 0.66 \\
\midrule
All-fake & 0.39 & 0.39 & - & 1.00 & 0.0 & 0.56 & - \\
All-real & 0.61 & - & 0.61 & 0.0 & 1.00 & - & 0.76 \\
\bottomrule
\end{tabular}%
\end{table}

%% file: table-satire.tex
\begin{table}[t!]%
\vspace{-2ex}%
\caption{Accuracies, class-wise precision, recall, and \fmeasure of predicting satire. Also shows the results for the best classifier of \newcite{rubin:2016}.}%
\label{table-satire}%
\scriptsize%
\centering%
\setlength{\tabcolsep}{5.25pt}%
\begin{tabular}{@{}l@{\hspace{10pt}}c@{\hspace{14pt}}cc@{\hspace{14pt}}cc@{\hspace{14pt}}cc@{}}
\toprule
\bfseries Features & \bfseries Accuracy & \multicolumn{2}{@{}c@{\hspace{14pt}}}{\bfseries Precision} & \multicolumn{2}{@{}c@{\hspace{14pt}}}{\bfseries Recall} & \multicolumn{2}{@{}c@{}}{\bfseries F$_{\mathbf{1}}$} \\
\cmidrule(r{14pt}){2-2}
\cmidrule(r{14pt}){3-4}
\cmidrule(r{14pt}){5-6}
\cmidrule{7-8}
& all & sat. & real & sat. & real & sat. & real \\
\midrule
Style & 0.82 & 0.84 & 0.80 & 0.78 & 0.85 & 0.81 & 0.82 \\
Topic & 0.77 & 0.78 & 0.75 & 0.74 & 0.79 & 0.76 & 0.77 \\
\midrule
All-sat. & 0.50 & 0.50 & - & 1.00 & 0.0 & 0.67 & - \\
All-real & 0.50 & - & 0.50 & 0.00 & 1.00 & - & 0.67 \\
\midrule
Rubin et al. & & 0.90 & & 0.84 & & 0.87 & \\
\bottomrule
\end{tabular}%
\end{table}

%% file: acl17-hyperpartisan-news-sum.tex
\section{Conclusion}

Fake news detection poses an interdisciplinary challenge: technology needs to be in place to extract factual statements from text, to match facts with a knowledge base, to dynamically retrieve and maintain knowledge bases from the web, to reliably assess the overall truthfulness of an article rather than individual statements, to do so in real time as news events unfold, to monitor the spread of fake news within and across social media, to measure the reputation of information sources, and to raise awareness in readers. These are perhaps only the most obvious things that can be done to tackle the problem, and as our cross-section of related work shows, many of them are tackled already. Given the recent hype around fake news, many more contributions will surface in the future---hopefully including our own.

Contemplating the task in its whole breadth, we sought ways to support these developments that are specific and can be reasonably addressed within the scope of a paper. Since many already attack fake news head on, by developing one way or another of fact-checking, we thought it worthwhile to mount our attack from another angle: writing style. Although difficult to be captured in a computer representation, it is just as difficult to be manipulated, let alone in a consistent way. While any knowledge base of sufficient size is prone to manipulation, authors of fake news may unwittingly leave traces of their hatred, political predisposition, or even disinterest, in case earning money is their only driving force, right within their fabrications.

In this paper, we have tried and succeeded to uncover at least some of these traces. News articles conveying a hyperpartisan world view can fairly easily be distinguished from more balanced news. Moreover, we have shown that the writing styles of otherwise opposing orientations, namely left-wing and right-wing, are in fact very similar: there appears to be a common style of extremism. We further show that satire can be distinguished well from other news, making at least sure that humor will not be outcast by fake news detection technology. Alas, we cannot claim to have solved fake news detection via style analysis alone.

\section*{Acknowledgements}

We thank Craig Silverman, Lauren Strapagiel, Hamza Shaban, Ellie Hall, and Jeremy Singer-Vine from BuzzFeed for making their data available, enabling our research.